\documentclass[preprint,review,5p,twocolumn]{elsarticle}

\usepackage{amssymb}
\usepackage{booktabs}
\usepackage{multirow}
\usepackage{array}
\usepackage{bbding}
\usepackage{graphicx}
\usepackage{lineno}
\usepackage{amsmath}
\newcommand{\tabincell}[2]{\begin{tabular}{@{}#1@{}}#2\end{tabular}}
\usepackage{color}
\newcommand{\red}[1]{\textcolor{black}{#1}}
\usepackage{url}
\usepackage{hyperref}
\RequirePackage[linesnumbered,ruled,vlined]{algorithm2e}

\journal{Arxiv}

\begin{document}

\begin{frontmatter}



\title{Exploring the Interactive Guidance for Unified and Effective Image Matting}


\author[1]{Dinghao~Yang\fnref{co-author}}
\author[1]{Bin~Wang\fnref{co-author}}
\author[2]{Weijia~Li\corref{corresponding}}
\ead{liweij29@mail.sysu.edu.cn}
\author[3]{Yiqi~Lin}
\author[1]{Conghui~He}

\affiliation[1]{organization={Shanghai Artificial Intelligence Laboratory},
            country={China}}
            
\affiliation[2]{organization={School of Geospatial Engineering and Science, Sun Yat-Sen University},
            country={China}}

\affiliation[3]{organization={AI Thrust, Information Hub, HKUST (Guangzhou)},
            country={China}}
            
\fntext[co-author]{D. Yang and B. Wang contribute equally to this work.}
\cortext[corresponding]{W. Li is the corresponding author.}

\begin{abstract}
Recent image matting studies are developing towards proposing trimap-free or interactive methods to complete the complex image matting task. Although avoiding the extensive labors of trimap interaction, existing methods still suffer from two limitations: (1) For the single image with multiple objects, it is essential to provide extra interaction information to help determining the matting target; (2) For transparent objects, the accurate regression of alpha matte from RGB image is much more difficult compared with the opaque ones. In this work, we propose UIM, a \textbf{U}nified \textbf{I}nteractive image \textbf{M}atting method, which solves the limitations and achieves satisfying matting results for any scenario. 
Specifically, UIM leverages multiple types of user interactions to avoid the ambiguity of multiple matting targets, and we compare the pros and cons of different user interaction types in detail. To unify the matting performance for transparent and opaque objects, we decouple image matting into two stages, \emph{i.e.}, foreground segmentation and transparency prediction. Moreover, \red{we propose a foreground consistency learning strategy to facilitate the feature extraction on mainstream synthetic matting dataset.} Experimental results demonstrate that UIM achieves \red{competitive} performance on Composition-1K test set and a synthetic unified dataset. Our code and models will be released at \href{https://github.com/Dinghow/UIM}{https://github.com/Dinghow/UIM}.
\end{abstract}




\end{frontmatter}


\section{Introduction}
%
%
%
%

\begin{figure}[t]
\begin{center}
\includegraphics[width=\linewidth]{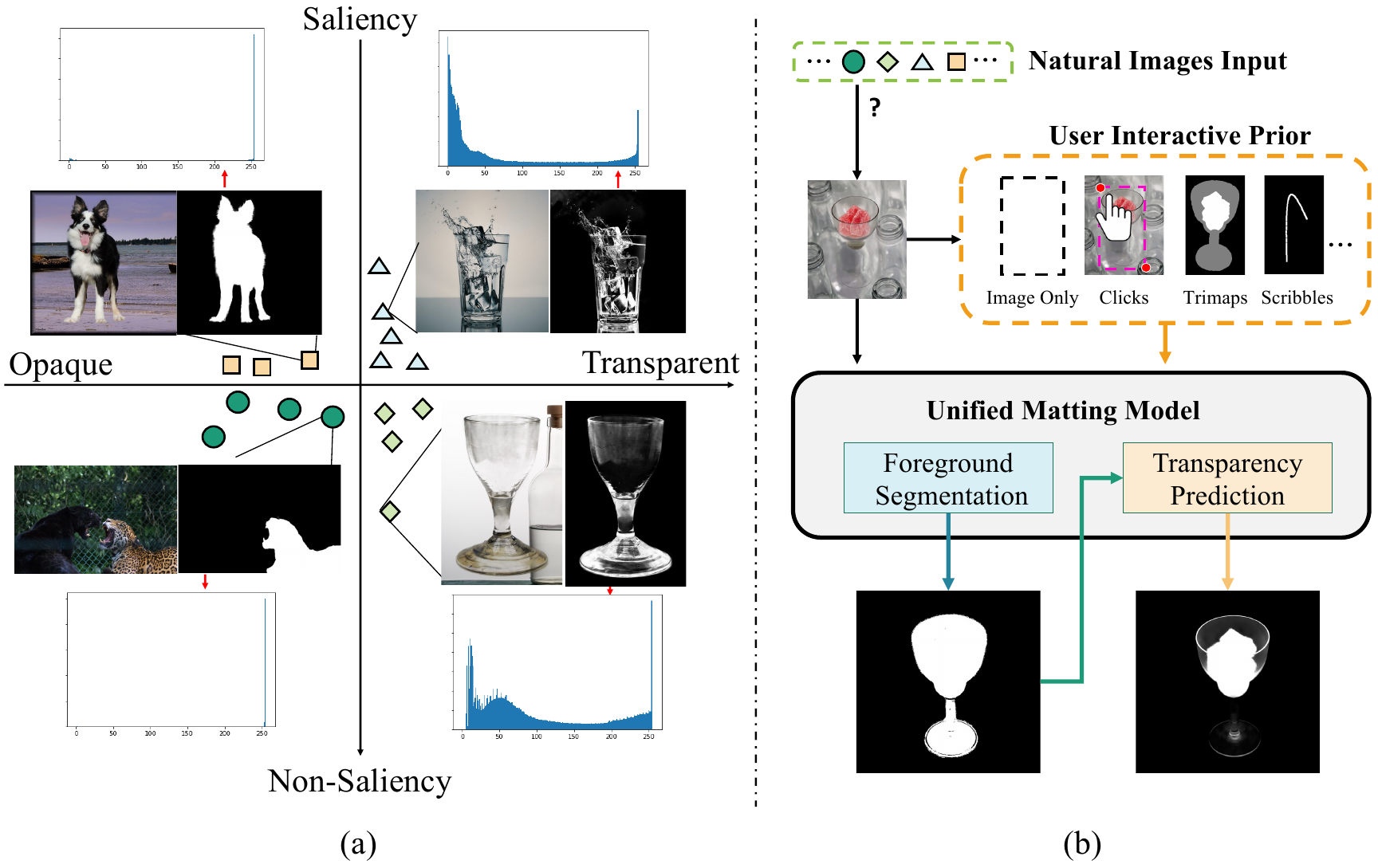}
\end{center}
\caption{(a): Images, alpha mattes and related foreground histogram statics of image matting data. The negative vertical semi-axis denotes samples from traditional matting datasets, of which the foreground object is salient and the background is clear. The positive vertical semi-axis denotes samples in real-world applications, in which there are multiple objects with more complex backgrounds. (b) An overview of the proposed Unified Interactive image Matting pipeline.}
\label{fig:uim_overview}
\end{figure}

Image matting is an important computer vision task with wide applications in image/video editing \cite{DIM, dong2020fusion, chen2021multi}, film production, virtual background for video conference and webcast \cite{BGMattingv1, BGMattingv2, MODNet}, etc. It aims at predicting the definite foreground, background and alpha matte of an image. Given an input image $I$, image matting can be formulated as:


\begin{equation}\label{matting}
    I_i = \alpha_i F_i + (1-\alpha_i)B_i, \quad \alpha_i \in [0, 1],
\end{equation}
where $F$, $B$ and $\alpha$ denote the foreground, background and alpha matte respectively. As $F_i$, $B_i$ and $\alpha_i$ at pixel $i$ in formula \ref{matting} are all unknown values, we can observe that image matting is a highly ill-posed problem. As shown in Figure \ref{fig:uim_overview}-(a), in real-world scenarios, the context of foreground and background can be various, especially in those images with multiple objects, and the distribution of alpha matte can be diversified as target objects can be opaque or transparent.

To address the ambiguity issue of the foreground target, and provide more prior information to help models solve the ill-posed problem, traditional image matting methods \cite{RobustMatting, GlobalMatting, Closed-FormMatting,KNNMatting, li2019parallel} and many mainstream deep learning based matting methods \cite{DIM, AdaMatting, GCAMatting, liu2021prior, park2022matteformer, fang2022user, hu2023effective} introduce trimaps as auxiliary inputs, \emph{i.e.}, coarse foreground, background and unknown region locations. However, obtaining trimaps suffers from time-consuming interaction labors and thus limits their applications.
To reduce these efforts, some recent studies propose trimap-free methods with only RGB image as input \cite{LFAMatting,qiao2020attention,AIM}. In fact, these methods still have difficulties in solving ambiguity of foreground target, which leads to uncertain prediction on images with multiple objects. Besides the trimap-based or trimap-free methods, \cite{zhou2023sampling, MGMatting} explores replacing the trimap with a segmentation mask generated from other segmentation models. \cite{BGMattingv1, BGMattingv2} adopts a background image instead of trimap to solve the ill-posed equation, while the background is hard to acquire since it must have the same intrinsic and extrinsic camera parameters with the input image.  

In addition, \cite{InteractiveMatting, DIIM} propose interactive matting methods using click interactions as guidance of foreground and background. However, it is hard to apply these methods to real-world image matting scenarios without considering the complex and diversified distribution of nature images. As shown in Figure \ref{fig:uim_overview}-(a), the foreground region histogram statistics of opaque and transparent objects can be dramatically different, as the alpha values of opaque one are mostly distributed around $0$ or $1$, while the alpha values of transparent one have a more even distribution. For the opaque cases, the matting task can be easily solved by a pixel-level classification model similar to the image segmentation task. However, it is difficult to use a unified model to deal with the opaque and transparent cases simultaneously as it requires the model to solve classification task and regression task at the same time. \red{Thus these interactive methods are easily confused on transparent objects, \emph{e.g.} clicking on the transparent foreground region may provide a similar prior as clicking on the background region.}
\red{As for other matting methods, several of them propose special design to solve the transparency distribution issue.} \cite{chen2018tom} treats this problem as refractive flow estimation, but the method is not unified for opaque cases and requires external rendering to generate ground truth. It is still necessary to handle the distribution discrepancy for building a matting model on real-world scenarios. 

Therefore, to address these two issues, we propose a unified interactive image matting method (UIM), which solves the above limitations and produces desirable matting effect for any type of image as shown in Figure \ref{fig:uim_overview}-(b).
First, to eliminate the ambiguity of matting targets for a single image with multiple objects, we introduce various types of user interaction information to image matting, \emph{e.g.} clicks with different strategies, scribbles. Furthermore, we classify the natural images into four classes as shown in Figure \ref{fig:uim_overview}-(a), and analyze these interactive manners on different scenarios in detail. \red{Compared with \cite{LFAMatting,AIM, qiao2020attention}, our proposed interaction manner can eliminate ambiguity without disturbing the information extraction capacity of the model.} 
Second, to unify the model capability for transparent and opaque objects, we decouple the image matting task into foreground segmentation and transparency prediction stages, which effectively improves the matting performance for transparent objects. The foreground segmentation stage focuses on the foreground/background classification problem, while the transparency prediction stage aims at solving the alpha matte regression problem in the foreground region.
To further improve the matting performance, we analyze the vagueness in the boundary region of the matting task and propose a multi-scale attentive fusion module to select local structure information and global context information automatically. Moreover, since most of the mainstream image matting dataset \cite{DIM, qiao2020attention} are synthetic, we also propose a general foreground consistency learning strategy for accelerating network convergence.

The main contributions of this work can be summarized as follows:
\begin{itemize}
\item We propose Unified Interactive image Matting, which firstly makes it possible for matting on all types of images. Results on multiple matting datasets show that our UIM achieves excellent performance compared with both trimap-based and trimap-free methods.
\item We introduce various types of user interactions to image matting task to eliminate the target ambiguity of a single image with multiple objects, and conduct a comprehensive comparison of these interaction manners.
\item We decouple the image matting task into foreground segmentation and transparency prediction stages, which improves the alpha matte regression accuracy and matting performance of transparent objects.
\item We propose a new training strategy for the synthetic image matting dataset, which utilizes the consistency of foregrounds and facilitates the capability of feature representation.
\end{itemize}

\section{Related Work}

\subsection{Traditional Methods}
As mentioned in the introduction section, alpha matte estimation directly with only RGB image is an ill-posed problem. Traditional hand-crafted methods usually limit the provision of valuable user input (e.g., trimaps \cite{BayesianMatting}, scribbles \cite{Scribbles}) to simplify the task. Essentially, these methods utilize association information of image patches with a similar appearance. 

Specifically, traditional image matting can be divided into sampling-based methods and propagation-based methods.  Sampling-based methods \cite{RobustMatting,GlobalMatting,zhong2013background,SharedSamplingMatting,SparseCodingMatting,BayesianMatting} collect definite foreground and background information to determine the candidate color of the true foreground and background, as well as the appropriate value of the corresponding alpha value for a given pixel. Due to the lack of considering the relationship between adjacent pixels in the unknown areas, these methods suffer from matting discontinuities and produce unreal prediction results. In contrast, the propagation-based methods can avoid this problem and usually obtain more robust predictions when dealing with complex images \cite{Closed-FormMatting,KNNMatting,FlowMatting}. In detail, this kind of method relies on the adjacent pixels' affinities and then propagates color information from the definite region into semi-transparent regions. 


\subsection{Deep Learning based Methods}
\noindent\textbf{Trimap-based image matting.} Benefiting from deep convolutional neural networks, the learning-based image matting method has gained rapid development in recent years. Deep Image Matting (DIM) \cite{DIM} proposes a trimap-based framework for image matting that can leverage high-level features and low-level features simultaneously. It also builds a large-scale image matting dataset, which is of great significance for the development of image matting. AdaMatting \cite{AdaMatting} points out the limitations of previous methods, which directly estimates the alpha matte from a coarse trimap and proposes a framework that decouples the image matting into trimap adaption and alpha estimation. Inspired by affinity-based methods and contextual attention, GCA-matting \cite{GCAMatting} designs a novel guided contextual attention module for image matting framework. MGMatting \cite{MGMatting} simplifies the trimap by computing a matte given an initial segmentation generated by existing methods, then processes the segmentation mask with image by progressive refinement network. Different from the above methods, our UIM can handle multiple types of interactive manners with an end-to-end network, providing a novel paradigm for matting on all types of images with better efficiency.

\hspace*{\fill} \\
\noindent\textbf{Trimap-free image matting.} Obtaining a suitable trimap itself is time-consuming and tedious, which can even take more than 10 minutes for some complicated cases \cite{InteractiveMatting}. To alleviate this problem, LFMatting \cite{LFAMatting} treats image matting as a foreground and background fusion task, taking only a single RGB image as input. However, the trimap-free image matting method is hard to get the desirable effect in real-world scenarios unless restricted to a specific domain (portrait matting \cite{MODNet,BGMattingv1,BGMattingv2}), due to the limited semantic representation of transparent foregrounds or non-salient foregrounds. To address the issue, AIM \cite{AIM} develops unified semantic representations for different types of the foreground and guide the matting network to focus on transition areas with an attention mechanism.

\begin{figure*}[t]
\begin{center}
\includegraphics[width=\linewidth]{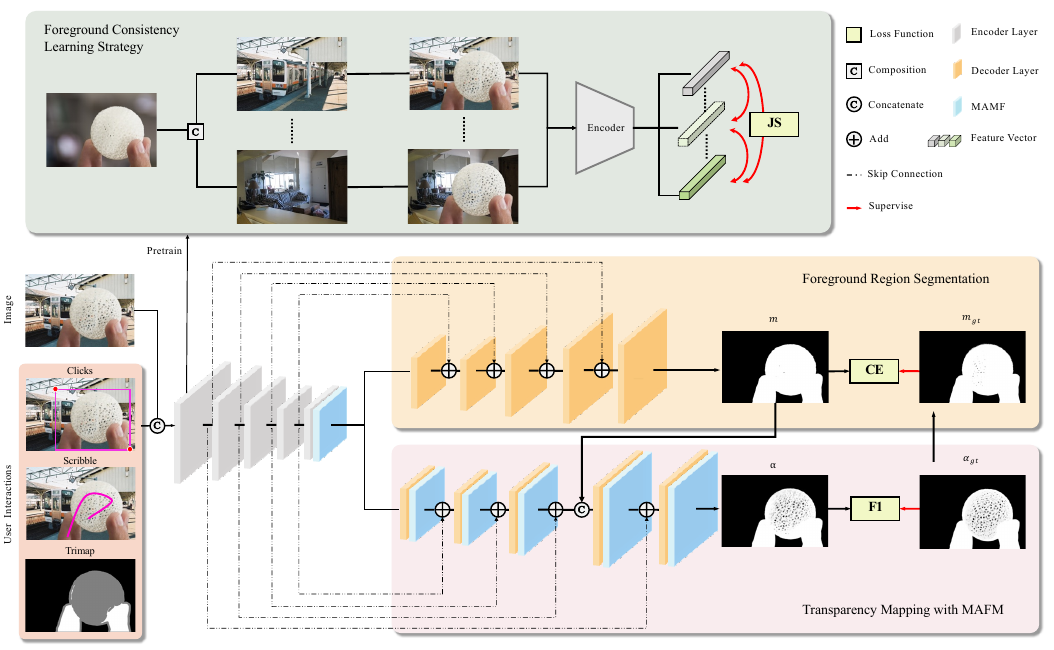}
\end{center}
\caption{Overview of the proposed UIM network. The ResNet-34 based encoder extract shared features of the input image and user interaction, then the segmentation decoder generates foreground region prediction, which is concatenated with features in the matting decoder and make final transparency mapping (TM). The user interactions of clicks include four types, \emph{i.e.}, foreground point, foreground and background points, bounding box and extreme points. The encoder is pretrained by foreground consistency (FC) learning strategy, and the multi-scale attentive fusion module (MAFM) is placed at the highest dimension after the encoder and each convolution layer after skip-connection of the matting decoder.}
\label{fig:main_network}
\end{figure*}

\subsection{Interactive Guidance}
Trimap-free image matting methods no longer need to spend extra time for drawing a trimap. However, the matting quality is still worse than trimap-based methods. The principal reason is that the matting model's distinction between foreground and background is ambiguous without the trimap guidance in some complicated scenarios (e.g. multiple objects or transparent objects).  Wei et al. \cite{InteractiveMatting} propose an interactive matting method to eliminate the ambiguity using click interactions when multiple objects appear in a single image. 
However, the click interaction fails to provide effective guidance when the matting targets are transparent, resulting in difficulties for applying the proposed method to complex real-world scenarios. \red{Ding et al. \cite{DIIM} propose a better clicking strategy for image matting, which defines the clicks into three types (\emph{i.e.}, positive, negative and transitive) as a brief representation of trimap region.
Nevertheless, this method still has performance limitations for transparent objects since it doesn't take the difference of distribution into consideration, and the clicking strategy requires more user interaction efforts compared with our UIM.} 

In terms of interaction methods, the bounding box, extreme points, and inside-outside are valuable user interactions that have driven the rapid evolution of interactive segmentation methods \cite{zhou2014interactive, dai2015boxsup,khoreva2017simple,jain2013predicting,lempitsky2009image,rother2004grabcut,xu2017deep,papadopoulos2017extreme,zhang2020interactive, tian2021interactive}. 
Different from the interactive segmentation task, the interactive matting is actually a soft segmentation task, and the target of interactive matting might be a transparent object. We analyze these interaction methods based on the above differences, and propose a unified framework which can produces desirable matting quality for all kinds of images.

\section{Method}

To enable matting on all types of images, we first disentangle and extend the image matting task with a unified paradigm. 
Then, based on the unified paradigm, we propose a unified interactive matting (UIM) network, as shown in Figure \ref{fig:main_network}.
UIM can utilize different kinds of user interactions as auxiliary input and perform precise matting prediction in two steps: 1) foreground region segmentation with an encoder-decoder architecture; 2) transparency mapping from segmentation for matting with a multi-scale attention mechanism. Additionally, we propose a novel foreground consistency learning strategy for the matting task, which gains better foreground feature extraction.

\subsection{Disentangle Unified Image Matting}


Inspired by \cite{AIM}, we classify the images into four classes in terms of the two key properties, \emph{i.e.} saliency and transparency: 1) salient opaque (SO), denoting the images with salient and opaque foreground objects, \emph{e.g.} salient portraits, animals and plants. As the definition of saliency is subjective in \cite{AIM}, here we explicate saliency as the clear information of which object to be matted, and vice versa; 2) salient transparent (ST), denoting the images with salient and transparent foreground objects, \emph{e.g.} glass, chiffon and plastics; 3) non-salient opaque (NSO), denoting the images with multi-objects that are opaque, \emph{e.g.} crowd and fauna; 4) non-salient transparent (NST), denoting the images with multi-objects that are transparent (\emph{e.g.} glasses, water drops and plastic bags).

Based on the above taxonomy, we first analyze the histogram statistics distribution of alpha values from all types of images. As shown in Figure \ref{fig:uim_overview}-(a), the alpha values of opaque objects (\emph{e.g.} portraits, animals) are mainly distributed in the absolute foreground region and absolute background region, which are very similar to the segmentation task \cite{chen2018tom}; for transparent objects, the histogram distributions are much more even.
This observation indicates that directly estimating the alpha matte from opaque and transparent objects through a unified model is very challenging due to the inherent distribution discrepancy.
To address this problem, we propose a two-step strategy, i.e., identifying foreground pixels and then estimating the alpha matte of foreground pixels. On the one hand, identifying foreground pixels can eliminate the distribution gap of predicting targets since regression targets from the two distributions both turn into a unified binary mask prediction. On the other hand, estimating the alpha matte of given foreground pixels can make the problem much easier compared with matting from raw images.
Nevertheless, matting on arbitrary images without the context of saliency is still a very ambiguous problem, especially for multi-object scenarios, since it is indispensable to indicate the saliency of which target needs to be matted.
In that case, auxiliary prior is necessary for coarsely locating the foreground and background region, which helps the model to better solve the over-undetermined problem. 

Finally, the unified image matting process can be decoupled into three sub-tasks: auxiliary prior by user interaction, foreground region segmentation, and Transparency Mapping with MAFM. Thus, unified image matting task can be formulated as:
\begin{equation}
    I = m \cdot t \cdot F + (1- m \cdot t) \cdot B, m \sim P_{m^{'}},
\end{equation}
where $m\in\{0,1\}$ refers to the foreground binary mask, $t\in [0,1]$ refers to the transparency, and $P_{m^{'}}$ is the region classification prior usually generated by handcraft interactions such as clicking or drawing. Then the optimization object is decoupled as foreground segmentation and transparency prediction of foreground region.

\subsection{User Interaction}

\noindent\textbf{Analysis on different types of user input.} There are two reasons why auxiliary user input is important for matting task. On the one hand, user input can provide the prior of foreground and background region to help model solve the \red{ill-posed problem better}. On the other hand, it helps to point out which object needs to be matted in multi-object non-salient cases. User can generate auxiliary input by clicking or drawing. 
Foreground point is the simplest user interaction type, which only requires a click on the foreground object. Scribble is another user interaction type that provides more foreground interaction via drawing lines. Both of the two aforementioned methods lack background information. Bounding box interaction uses two clicks (left-top and right-bottom) to indicate a coarse foreground region, while the pixels out of it belong to certain background. Moreover, extreme points interaction \cite{maninis2018deep} provides both foreground and background priors from borderline points (top, bottom, left-most, right-most of object) using a convenient interaction manner. In addition, foreground and background points interaction \cite{zhang2020interactive} can provide more certain priors via clicking on absolute regions. However, clicking on foreground region is erratic and may lead to serious errors due to the transparency.
Besides, trimap segment the image into foreground, background and unknown regions by manual drawing, which is the most widely-used user interaction type for image matting but surfers from heavy workload.

\hspace*{\fill} \\
\noindent\textbf{Simulated user input.} 
In our UIM, the choice of user interaction types is flexible according to the actual requirements. All of the available user interactions are concatenated with the RGB image as the network input. Similar to interactive segmentation methods \cite{maninis2018deep, zhang2020interactive}, each type of user interaction \red{in training and testing stages} is generated via the following simulation methods.
(1) For the foreground point, we sample the furthest point from the alpha boundary with a random perturbation of \red{0$\sim$5 pixels}. 
(2) For the bounding box, we expand the ground truth alpha by \red{10 pixels} and fill in the rectangle region with the value of 0, which provides more region prior than the point type. 
(3) For the background points, we select four corner points from the bounding box. (4) For extreme points, we follow the simulation method in \cite{maninis2018deep}, \emph{i.e.}, jittering the top, bottom, left-most and right-most points from alpha. (5) For scribble, we first randomly select 3 or 4 control points from the alpha region. Then we generate 5 B-spline curves with interpolation and dilation, and select the curve that covers the largest area of alpha. (6) For trimap, we follow \cite{DIM}, \emph{i.e.}, dilating randomly from alpha to generate foreground and background, and the remains are marked as unknown.


\subsection{Foreground Region Segmentation}
To decouple image matting task, we design a dual decoder architecture with a shared feature encoder, as shown in Figure \ref{fig:main_network}. The encoder of the network dedicates to extract multi-scale features from the images and interactions, thus UIM adopts a shared encoder to execute primary feature embedding operation, and the segmentation and matting decoders are designed for different sub-tasks with special architecture. The top branch is the mask decoder. We use the popular U-Net \cite{UNet} architecture to solve the segmentation task, which is a broadly used architecture with satisfying performance in previous image matting methods \cite{GCAMatting, MGMatting, AIM}. The ground truth mask $m_{gt}$ is generated from ground truth alpha $\alpha_{gt}$ with a threshold $t$,

\begin{equation}
    m_{gt}(i,j) = 
\begin{cases}
    1& \text{$\alpha_{gt}(i,j)>t$}\\
    0& \text{else}
\end{cases}.
\end{equation}
where $(i, j) \in \mathbb{R}^{h \times w}$. In general, we set $t$ as $0$ in this paper, and this branch can predict full foreground region mask.

\subsection{Transparency Mapping with MAFM}
As shown in Figure \ref{fig:main_network}, the bottom branch is the matting decoder using the same architecture as the mask decoder. To generate precise alpha matte, we introduce a multi-scale attentive fusion and transparency mapping method.

\begin{figure}[h]
\begin{center}
\includegraphics[width=0.95\linewidth]{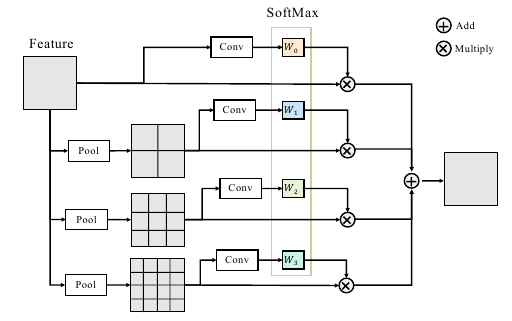}
\end{center}
\caption{The detailed architecture of multi-scale attentive fusion module (MAFM).}
\label{fig:mafm}
\end{figure}

\noindent\textbf{Multi-scale attentive fusion.} 
\red{Manual interactions solve the ambiguity of multiple objects, making the network focus on the object area. However, we find the color and texture information in object areas are usually complicated, and the interlaced foreground and background (\emph{e.g.} samples in Figure. \ref{fig:mafm}) bring challenges to network prediction.} To facilitate the network to process objects with complex boundaries, adaptive feature extraction is helpful (\emph{e.g.} low-frequency features for the details of objects, and high-frequency features for the main shape of objects). As shown in Figure \ref{fig:mafm}, inspired by \cite{PSPNet, liu2019learning}, we \red{introduce a multi-scale attentive fusion module (MAFM) to image matting task}, which is a plug and play module for feature representation. First, the input feature $x_0$ is down-sampled to $x_1, x_2, x_3$ by avg-pooling with the kernel size of $1 \times 1, 3 \times 3$ and $5 \times 5$, respectively. Then the features are embedded by convolution layers, and the weight of different scale $w_i$ is calculated as follow:

\begin{equation}
    w_i = \frac{\mathcal{C}(x_{i \rightarrow 0})}{\sum\limits_{j = 0}^{3}{\mathcal{C}(x_{j \rightarrow 0})}},
\end{equation}
where $\mathcal{C}$ refers to convolution and activation operation, and the right arrow refers to interpolate to the same size. Then the final fusion feature $y$ is acquired by:
\begin{equation}
    y = \sum\limits_{i = 0}^{3}w_i \cdot x_{i \rightarrow 0}.
\end{equation}

\hspace*{\fill} \\
\noindent\textbf{Transparency mapping.} The predicted foreground mask $m$ is first resized to the same shape as low-level features $l$ from the matting decoder and then concatenated. With the supervision from $\alpha_{gt}$, a pixel-level transparency mapping $\mathcal{M}: (m || l) \rightarrow \alpha$ is established as:

\begin{equation}
    \alpha_{k}  = \mathcal{M}_{\Theta}( m_{k} || l_{k}),
\end{equation}
where $\alpha$ is the predicted alpha matte, k refers to the index of pixel, and $\mathcal{M}_{\Theta}$ is a function to fit the mapping. Here we implement it using two convolution layers.

\subsection{Training Strategy}
\noindent\textbf{Loss functions.} For the mask prediction, we adopt cross-entropy loss:

\begin{equation}
     \mathcal{L}_{ce} = -(m_{gt}log(m) + (1-m_{gt})log(1-m)).
\end{equation}

For the final alpha matte prediction, we adopt $l_1$ regression loss:

\begin{equation}
     \mathcal{L}_{l_1} = |\alpha_{gt} - \alpha|,
\end{equation}

The final loss function is:

\begin{equation}
     \mathcal{L}_{final} = \mathcal{L}_{ce} + \lambda \mathcal{L}_{l_1},
\end{equation}
where $\lambda$ is a balancing parameter (here we set as 1).

\hspace*{\fill} \\
\noindent\textbf{Foreground Consistency Learning Strategy.} The label of alpha matte is generated by triangulation \cite{AlphaMatting} or photoshop \cite{DIM}. Both of the processes require a large expenditure of time and effort. Thus current mainstream matting datasets \cite{DIM, qiao2020attention} are all synthetic, using limited unique foregrounds with one-to-many new backgrounds to composite a considerable amount of data. For instance, for the Composition-1K dataset, every foreground are composed with 10 backgrounds from PASCAL VOC \cite{everingham2010pascal} and MS COCO \cite{lin2014microsoft}. Although other data augmentation methods \cite{GCAMatting} are adopted to make the synthetic data more diverse, there is still a lot of feature redundancy, resulting in the waste of computing resources and inefficient network convergence. Dedicating to overcoming these limitations, we propose a novel training strategy on the synthetic dataset. \red{Let $f_i$ denotes a foreground of $F \in \mathbb{R}^{M}$, $B_i = \{b_{i1}, b_{i2}, ..., b_{ij}\}$ denotes the related background set with $N$ samples, and $S_i = \{s_{i1}, s_{i2}, ..., s_{ij}\}$ is the composited image set. We introduce Jensen-Shannon divergence (JS), a symmetric distribution similarity measurement metric to calculate the difference among features $E$ extracted from $S$:}

\begin{equation}
    \mathcal{L}_{cons} = \frac{MN(N-1)}{2}\sum\limits_{i=1}^{M}\sum\limits_{j=1}^{N}\sum\limits_{k=i+1}^{N} JS(e_{ij}, e_{ik}),
\end{equation}
where $e_{ij}$ and $e_{ik}$ are the output features of the shared encoder. Since only the background regions are different, \red{with the consistency constraint of symmetric loss function}, the encoder can better learn foreground feature representation. We use this strategy to pre-train an encoder with strong representation capacity for the further matting task with $\mathcal{L}_{final}$.

\begin{figure*}[t]
\begin{center}
\includegraphics[width=\linewidth]{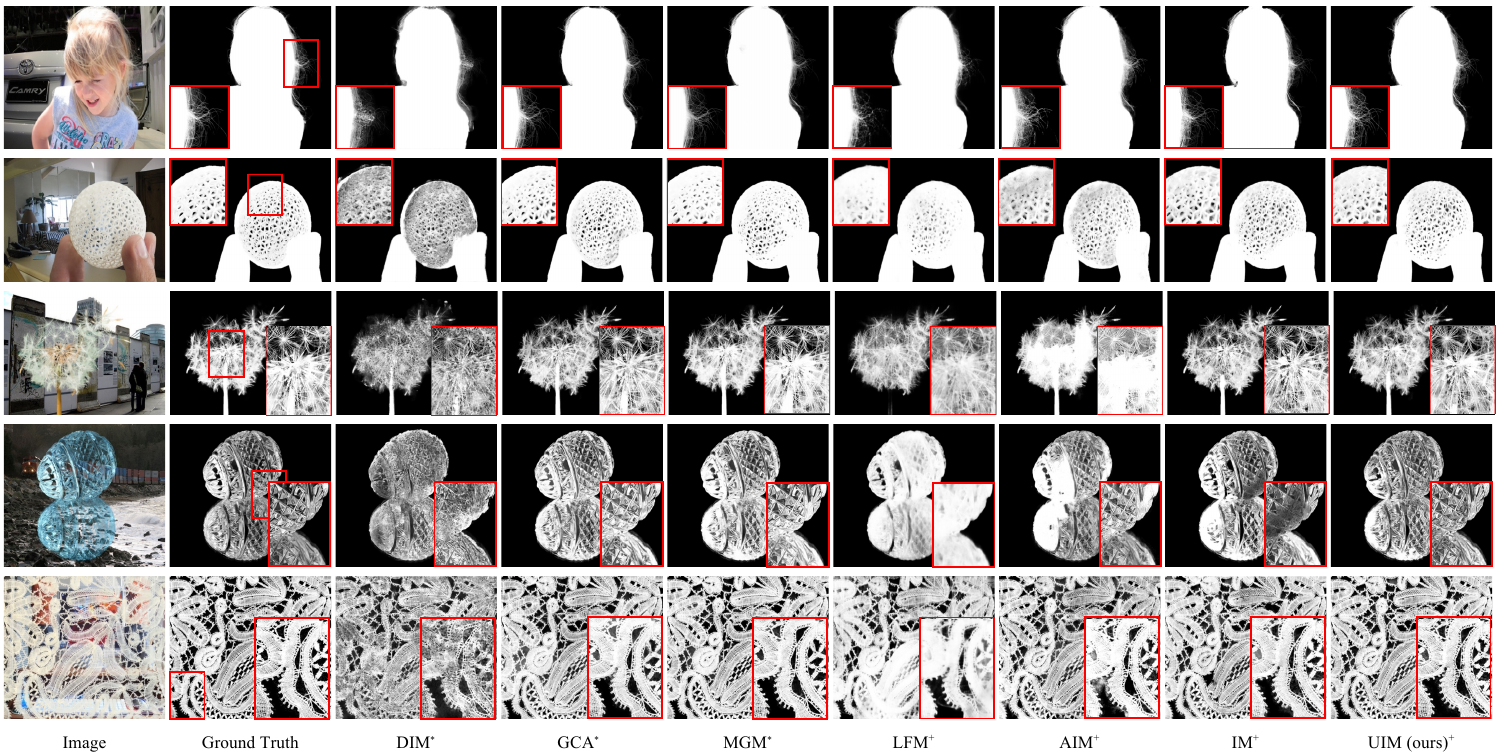}
\end{center}
\caption{The visual comparison results on Composition-1K test set. The results of UIM are generated with bounding box interaction. $^*$ denotes trimap-based methods, and $^+$ denotes trimap-free methods.}
\label{fig:comparison_with_sota_comp1k}
\end{figure*}

\section{Experiments}
\subsection{Dataset and Evaluation Protocols}

\noindent\textbf{Dataset.} We train and test our model on the widely-used synthetic Composition-1K dataset \cite{DIM}, which includes 43,100 training samples and 1,000 testing samples, with 431 and 50 unique foreground images, respectively. The ratio of different categories of images in the test set is 31 : 19 : 0 : 0 for  SO, ST, NSO and NST. To make a more balanced test set, we create a synthetic unified test dataset based on Composition-1K. First, we select SO and ST images from free-license websites as background, then we composite them with SO and ST foreground from Composition-1K to generate multi-object non-salient samples, as shown in Figure \ref{fig:uim_overview}-(a). Finally, the new test dataset contains 805 samples, with SO : ST : NSO : NST = 310 : 140 : 280 : 75.

\hspace*{\fill} \\
\noindent\textbf{Trimap-based metrics and trimap-free metrics.} The matting results are usually evaluated \cite{DIM} by Mean Squared Error (MSE), Sum of Absolute Differences (SAD), Gradient (Grad.) and Connectivity (Conn.). Previous trimap-based methods \cite{DIM, AdaMatting, GCAMatting} only compute loss and metrics in unknown region. Meanwhile, there are a few trimap-free methods with a clear statement of the metric region. \cite{BGMattingv2} generates a trimap from ground truth alpha and only concerns the unknown region, while \cite{LFAMatting} sums the metrics over the entire image. The fuzzy definition of metric region may lead to unfair comparison (\emph{e.g.} lower MSE), thus we use two types of metrics in the following experiments with clear indication, \emph{i.e.} the trimap-based metrics that computes on the unknown region, and trimap-free metrics that computes overall pixels of the image region.

\subsection{Implementation Details}
\label{sec:detail}

\noindent\textbf{Network architectures.} The backbone of the proposed dual-decoder network is ResNet-34 \cite{ResNet}, and both decoders have skip-connection with the shared encoder at each stage. The channel number of the first convolution layer is adjustable. Except $3$ channels for the RGB image, the extra channels are flexible for different interactive inputs. To adaptively select features, the MAFM block is placed at the highest dimension after encoder and each convolution layer after skip-connection of the matting decoder.

\hspace*{\fill} \\
\noindent\textbf{Training details.} First, we pre-train the encoder using the proposed foreground consistency training strategy. We crop the input image and user interaction to $512 \times 512$ pixels, which are augmented by random jitter. Then we train the whole network in an end-to-end manner with the pre-trained encoder. To alleviate the synthetic redundancy and improve the generalization ability, we augment the training data following \cite{GCAMatting}, including random combination, random affine transformation, random crop and random jitter. We train our model on 4 Tesla V100, using adam optimizer with $\beta_1 = 0.5$ and $\beta_2 = 0.999$, and a learning rate of $4 \times 10^{-4}$. For the pre-training stage, the policy is \emph{poly} with a power of 0.9, and a batch size of 160 for 20 epochs; For the main training stage, the policy is \emph{warm-up} with 5,000 iterations increment and a cosine learning rate scheduler is adopted, with a batch size of 64 for 120 epochs.

\subsection{Comparison with State-of-the-art Methods}

\begin{figure*}[t]
\begin{center}
\includegraphics[width=0.8\linewidth]{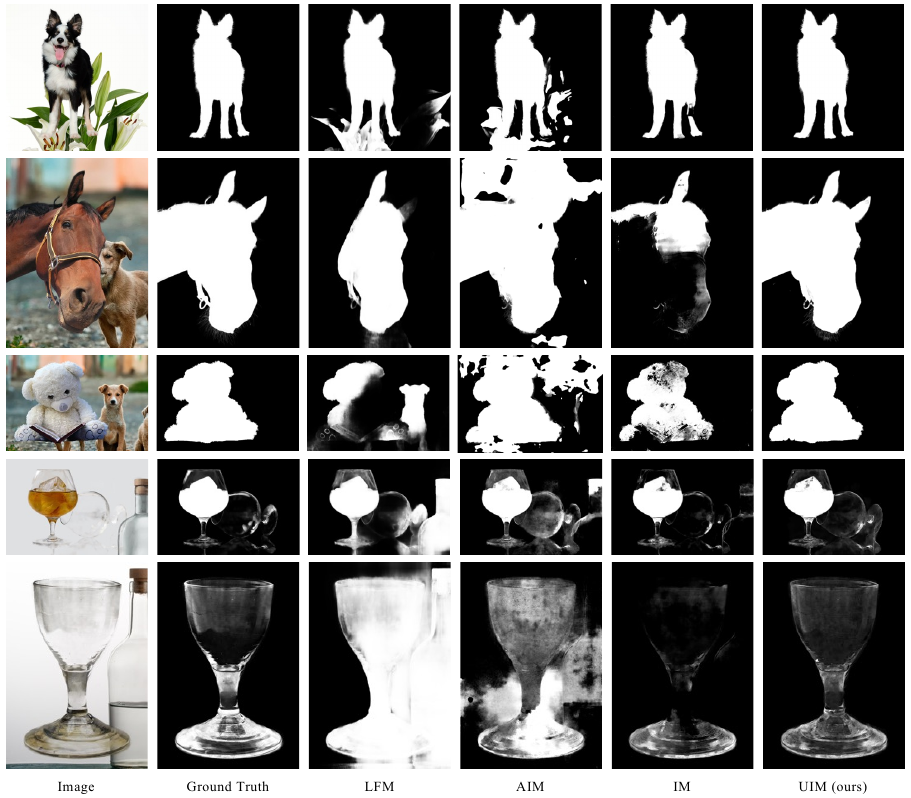}
\end{center}
   \caption{The visual comparison results on the synthetic unified dataset. The results of UIM are generated with bounding box interaction.}
\label{fig:comparison_with_sota_unified}
\end{figure*}

\noindent\textbf{Evaluation on the Composition-1K dataset.} We compare our UIM with the state-of-the-art image matting methods on Composition-1K test set \cite{DIM}, which is composed of 50 foreground images with 1,000 background images from PASCAL VOC\cite{VOC}.
As there are few trimap-free methods with officially released code \footnote{The released model of Late Fusion Matting is trained with additional data, thus the results are only for reference.}, we select AIMNet \cite{AIM} with the best performance for re-implementation following the experimental settings in its papers and released code. Meanwhile, we also reproduce the Interactive Matting method \cite{InteractiveMatting} for comparison.
To guarantee a fair comparison and vindicate the effectiveness of UIM architecture, we also adopt trimap as auxiliary input when we compare the performance of UIM with existing trimap-based methods.
As illustrated in Table \ref{tab:comparison_with_sota_comp1k}, UIM achieves competitive results among trimap-based methods and outperforms all state-of-the-art methods in trimap-free categories. 
The qualitative results are depicted in Figure \ref{fig:comparison_with_sota_comp1k}. Our UIM only adopts bounding box as auxiliary input, which can be generated by two clicks, achieving comparable results as trimap-based methods.

 \begin{table}[h]
    \tiny
	\centering
	\caption{The quantitative evaluation results on Composition-1K test set, in terms of trimap-based evaluation metrics. UI is the abbreviation of user interaction, fg/bg denotes foreground and background points, and bbox denotes the bounding box.}	
	\begin{tabular}{l|c|cccc} 
	\toprule
     Methods & UI. & MSE & SAD & Grad. & Conn.\\
	\midrule
    Learning Based Matting \cite{LearningBasedMatting} & trimap & 0.048 & 113.9	& 91.6 & 122.2\\
    Closed-Form Matting	\cite{Closed-FormMatting} & trimap & 0.091 & 168.1 & 126.9 & 167.9\\
    KNN Matting \cite{KNNMatting} & trimap & 0.103 & 175.4 & 124.1 & 176.4\\
    Deep Image Matting \cite{DIM} & trimap & 0.014 & 50.4 & 31.0 & 50.8\\
    AdaMatting \cite{AdaMatting} & trimap & 0.010 & 41.7 & 16.9 & -\\
    GCA Matting\cite{GCAMatting} & trimap & 0.009 & 35.3 & 16.9 & 32.5\\
    Context-Aware Matting \cite{Context-AwareMatting} & trimap & 0.008 & 35.8 & 17.3 & 33.2\\
    MGMatting \cite{MGMatting} & trimap & 0.007 & 32.1 & 14.0 & 27.9\\
    MatteFormer \cite{park2022matteformer} & trimap & \textbf{0.004} & \textbf{23.8} & \textbf{8.7} & \textbf{18.9}\\
    UIM (ours) & trimap & 0.007 & 31.4 & 13.5 & 26.8\\
    \midrule
    Late Fusion Matting \cite{LFAMatting} & N.A. & 0.039 & 88.4 & 83.5 & 71.7\\
    AIMNet \cite{AIM} & N.A. & 0.027 & 65.8 & 44.7 & 53.7\\
    Interactive Matting \cite{InteractiveMatting}	& fg/bg & 0.042 & 73.4 & 48.6 & 54.9\\
    UIM (ours) & bbox & \textbf{0.012} & \textbf{38.2} & \textbf{17.9} & \textbf{33.8}\\
	\bottomrule
	\end{tabular}
	\label{tab:comparison_with_sota_comp1k}
\end{table}
 \begin{table}[h]
    \footnotesize
	\centering
	\caption{The quantitative evaluation results on the synthetic unified dataset, in terms of trimap-free evaluation metrics. The results of UIM are test with bounding box interaction.}	
	\begin{tabular}{l|c|cccc} 
	\toprule
     Methods & Categories & MSE & SAD & Grad. & Conn.\\
	\midrule
    \multirow{5}{*}{\tabincell{l}{Late Fusion \\ Matting \cite{LFAMatting}}} & SO & 0.005 & 32.3 & 24.0 & 22.5 \\
    & ST & 0.022 & 118.8 & 94.8 & 86.5\\
    & NSO & 0.097 & 272.2 & 47.5 & 70.5\\
    & NST & 0.092 & 336.2 & 160.1 & 134.5\\
    & overall & 0.048 & 158.8 & 57.2 & 60.8\\
	\midrule
    \multirow{5}{*}{AIMNet \cite{AIM}} & SO & 0.005 & 32.4 & 19.3 & 24.8\\
    & ST & 0.014 & 96.1 & 50.9 & 64.6\\
    & NSO & 0.212 & 546.8 & 132.7 & 133.7\\
    & NST & 0.185 & 572.0 & 163.5 & 111.7\\
    & overall & 0.095 & 272.7 & 77.7 & 77.7\\
	\midrule
    \multirow{5}{*}{\tabincell{l}{Interactive \\ Matting \cite{InteractiveMatting}}} & SO & 0.004 & 27.2 & 17.3 & 23.6 \\
    & ST & 0.010 & 66.7 & 32.0 & 52.7\\
    & NSO & 0.089 & 210.6 & 77.8 & 89.0\\
    & NST & 0.129 & 143.0 & 95.1 & 65.7\\
    & overall & 0.083 & 214.23 & 68.3 & 65.8 \\
	\midrule
    \multirow{5}{*}{UIM (ours)} & SO & 0.002 & 18.9 & 10.0 & 16.1\\
    & ST & 0.006 & 56.3 & 26.8 & 47.9\\
    & NSO & 0.065 & 190.4 & 55.0 & 75.0\\
    & NST & 0.088 & 288.5 & 88.5 & 88.1\\
    & overall & \textbf{0.033} & \textbf{110.2} & \textbf{35.9} & \textbf{48.8}\\
	\bottomrule
	\end{tabular}
	\label{tab:comparison_with_sota_unified}
\end{table}

\hspace*{\fill} \\
\noindent\textbf{Evaluation on the synthetic unified dataset.} Previous matting methods are trained and tested on the synthetic datasets with simple backgrounds and salient foregrounds, thus it is uncertain whether these methods are capable enough for all types of images. As mentioned before, saliency and transparency are two key factors for this problem, thus we generate a synthetic unified dataset with images from four categories. We further evaluate our UIM and other state-of-the-art trimap-free methods on this dataset. The evaluation results and some representative samples are demonstrated in Table \ref{tab:comparison_with_sota_unified} and Figure \ref{fig:comparison_with_sota_unified}. 
As shown in Table \ref{tab:comparison_with_sota_unified}, our UIM achieves the best performance on all categories, and makes a significant improvement compared with other trimap-free methods. Late Fusion Matting \cite{LFAMatting} and AIMNet \cite{AIM} only utilize RGB images as input, which perform well on SO and ST images. However, the performance drops seriously for non-salient images since there is no prior information to demonstrate the foreground saliency in multi-object cases. Interactive Matting (IM) \cite{InteractiveMatting} uses foreground and background points as prior information, which obtains more robust performance on NSO images. Nevertheless, due to the transparency of foreground objects, it is not easy for the network to determine whether the point is a hint of foreground or background, resulting in the failure of IM \cite{InteractiveMatting} on NST images. As shown in Figure \ref{fig:comparison_with_sota_unified}, for LFMatting, AIMNet and IM, the details of matting predictions are coarse and error-prone. In contrast, UIM only adopts the bounding box as auxiliary prior information, which provides both enough context of saliency and strong robustness of transparent foregrounds, generating the image matting results with much finer details. The obvious qualitative and quantitative improvements of UIM also come from the better foreground object recognition capability by foreground consistency learning strategy (FC), easier optimization convergence via disentangling the matting problem (TM), and robust and precise feature representation using multi-scale design (MAFM), which will be analyzed in the following section.

\hspace*{\fill} \\
\noindent\textbf{Effectiveness of UIM network.} \red{To further validate the effectiveness of UIM network architecture, we compare our UIM with MGMatting \cite{MGMatting} (trimap-based) and AIMNet \cite{AIM} (trimap-free) using the same user interactions. As shown in Table. \ref{tab:comparison_with_mg_aim}, UIM outperforms the other two methods for all types of interactions, which vindicates that the performance improvement of UIM comes from the network design instead of the extra priors provided by user interactions.}

\begin{table}[h]
    \tiny
    \centering
    \caption{Comparison on Composition-1K. The metrics are trimap-based. \dag denotes the mask are generated by UIM (bbox).}	
    \begin{tabular}{l|l|cccc} 
    \toprule
     Dataset & Interactions & MSE & SAD & Grad. & Conn.\\
    \midrule
    \multirow{5}{*}{MGMatting \cite{MGMatting}} & FG point & 0.093 & 131.6 & 96.5 & 82.6 \\
    & FG/BG points & 0.037 & 75.2 & 58.5 & 59.7\\
    & Extreme points & 0.023 & 54.2 & 31.2 & 48.9\\
    & Scribble & 0.054 & 85.5 & 51.9 & 50.8\\
    & Bounding box & 0.013 & 46.4 & 23.5 & 40.8\\
    & Trimap & \textbf{0.007} & 32.1 & 14.0 & 27.9\\
    & Mask \dag & 0.010 & 38.3 & 18.1 & 33.2\\
    \midrule
    \multirow{5}{*}{AIMNet \cite{AIM}} & FG point & 0.120 & 152.6 & 103.0 & 97.8 \\
    & FG/BG points & 0.041 & 88.2 & 64.3 & 61.9\\
    & Extreme points & 0.024 & 61.3 & 34.4 & 51.0\\
    & Scribble & 0.062 & 91.6 & 60.4 & 59.7\\
    & Bounding box & 0.018 & 51.5 & 32.5 & 56.8\\
    & Trimap & 0.027 & 65.8 & 44.7 & 53.7\\
    & Mask \dag & 0.024 & 63.2 & 41.5 & 49.6\\
    \midrule
    \multirow{5}{*}{UIM(ours)} & FG point & 0.087 & 123.9 & 85.2 & 74.2 \\
    & FG/BG points & 0.032 & 69.0 & 51.7 & 50.4\\
    & Extreme points & 0.018 & 51.2 & 28.9 & 42.2\\
    & Scribble & 0.046 & 72.8 & 42.0 & 43.2\\
    & Bounding box & 0.012 & 38.2 & 17.9 & 33.8\\
    & Trimap & \textbf{0.007} & 31.4 & 13.5 & \textbf{26.8}\\
    & Mask \dag & \textbf{0.007} & \textbf{30.9} & \textbf{13.1} & 26.9\\
    \bottomrule
    \end{tabular}
    \label{tab:comparison_with_mg_aim}
\end{table}

\subsection{Ablation Studies}
We conduct extensive experiments to verify the effectiveness of the proposed unified interactive image matting pipeline and related UIM network. First, we compare different user interaction types in detail, and summarize the advantages and disadvantages of each interaction type. Then we quantitatively analyze each module of UIM.

\begin{figure}[h]
\begin{center}
\includegraphics[width=\linewidth]{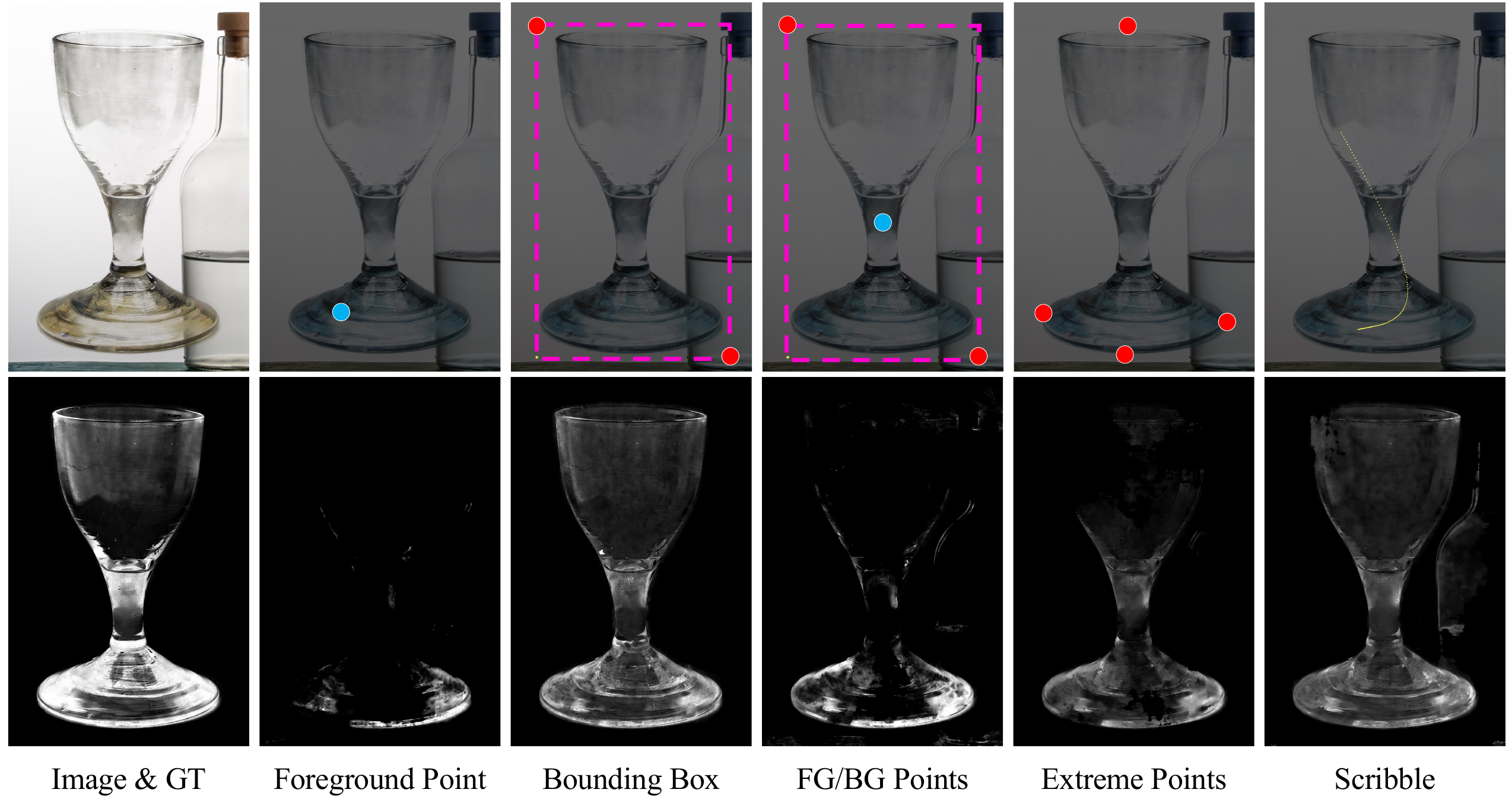}
\end{center}
\caption{The comparison of user interaction types. The top row illustrates the interaction types, and the bottom row is the prediction result.}
\label{fig:annotation_comparison}
\end{figure}

 \begin{table*}[t]
    \tiny
	\centering
	\caption{User interaction ablation studies on the synthetic unified dataset. The metrics are trimap-free.}
	\begin{tabular}{l|cccc|cccc|cccc|cccc} 
	\toprule
	\multirow{2}{*}{Innotation} & \multicolumn{4}{c|}{SO} & \multicolumn{4}{c|}{ST} & \multicolumn{4}{c|}{NSO} & \multicolumn{4}{c}{NST}\\ 
      & MSE & SAD & Grad. & Conn. & MSE & SAD & Grad. & Conn. & MSE & SAD & Grad. & Conn. & MSE & SAD & Grad. & Conn.\\
     \midrule
     FG point & 0.074 & 257.0 & 62.1 & 139.1 & 0.046 & 149.5 & 59.9 & 95.6 & 0.251 & 686.9 & 109.9 & 99.1 & 0.105 & 342.5 & 155.6 & \textbf{73.9}\\ 
     FG/BG points & 0.052 & 176.3 & 51.5 & 88.6 & 0.010 & 70.8 & 32.2 & 52.2 & 0.143 & 442.9 & 129.1 & 170.7 & 0.093 & 311.4 & 126.9 & 82.3\\ 
     Extreme points & 0.012 & 51.6 & 15.8 & 37.8 & \textbf{0.006} & \textbf{55.7} & 26.0 & \textbf{47.2} & 0.125 & 360.6 & 74.1 & 119.6 & 0.096 & 318.5 & 102.6 & 90.3\\ 
     Scribble & 0.014 & 58.5 & \textbf{9.1} & 42.0 & \textbf{0.006} & 56.0 & \textbf{25.5} & 48.5 & 0.137 & 371.6 & 63.9 & 125.2 & 0.098 & 315.5 & 106.3 & 95.5\\
     Bounding box & \textbf{0.002} & \textbf{18.9} & 10.0 & \textbf{16.1} & \textbf{0.006} & 56.3 & 26.8 & 47.9 & \textbf{0.065} & \textbf{190.4} & \textbf{55.0} & \textbf{75.0} & \textbf{0.088} & \textbf{288.5} & \textbf{88.5} & 88.1\\ 
	\bottomrule
	\end{tabular}
	\label{tab:interactive_ablation_4classes}
\end{table*}

\begin{figure*}[htbp]
\begin{center}
\includegraphics[width=\linewidth]{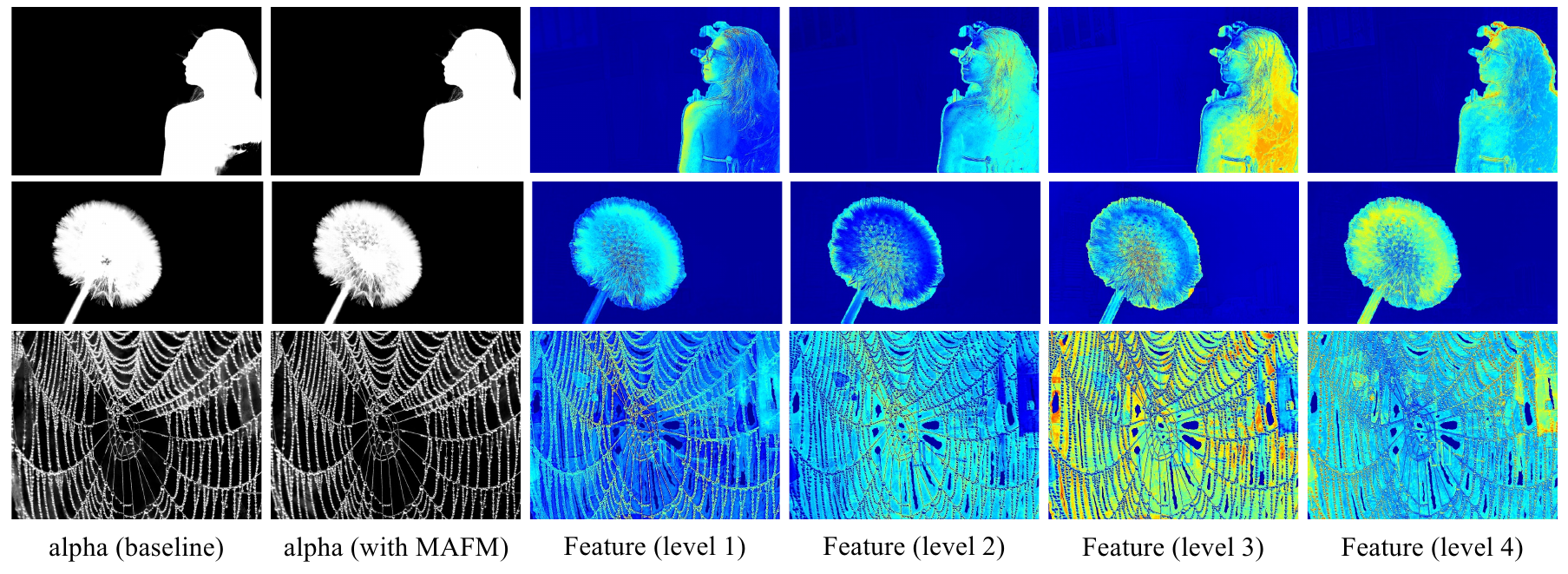}
\end{center}
  \caption{Visualization of feature maps from multi-scale attentive fusion module (MAFM). Feature maps at level 1-3 denote the features with $1 \times 1$, $3 \times 3$, $5 \times 5$ pooling, respectively. And level 4 denotes feature map processed from input features.}
\label{fig:mafm}
\end{figure*}

\begin{table}[h]
    \tiny
	\centering
	\caption{User interaction ablation studies on Composition-1K test set and the synthetic unified dataset. The metrics are trimap-free.}	
	\begin{tabular}{l|l|cccc} 
	\toprule
     Dataset & Interactions & MSE & SAD & Grad. & Conn.\\
	\midrule
    \multirow{5}{*}{Composition-1K} & FG point & 0.077 & 265.9 & 103.8 & 142.9 \\
    & FG/BG points & 0.042 & 165.9 & 74.9 & 92.2\\
    & Extreme points & 0.015 & 77.3 & 33.1 & 59.9\\
    & Scribble & 0.039 & 139.0 & 39.6 & 58.3\\
    & Bounding box & \textbf{0.006} & \textbf{49.9} & \textbf{25.2} & \textbf{43.6}\\
    \midrule
    \multirow{5}{*}{\tabincell{l}{synthetic unified \\ dataset}} & FG point & 0.133 & 395.8 & 87.0 & 111.6 \\
    & FG/BG points & 0.080 & 263.3 & 82.2 & 113.4\\
    & Extreme points & 0.058 & 184.7 & 46.0 & 72.8\\
    & Scribble & 0.064 & 190.9 & 40.1 & 77.1\\
    & Bounding box & \textbf{0.033} & \textbf{110.2} & \textbf{35.9} & \textbf{48.8}\\
	\bottomrule
	\end{tabular}
	\label{tab:interactive_ablation_overall}
\end{table}

\hspace*{\fill} \\
\noindent\textbf{User interaction.} Figure \ref{fig:annotation_comparison} illustrates the types of user interaction. The time cost of each interaction type is counted manually following \cite{bearman2016s}. One click takes 2.4s, and one scribble takes 10.9s on average. The quantitative comparison results of different user interactions are shown in Table \ref{tab:interactive_ablation_overall}, from which we can see that the bounding box outperforms the other interaction types on the Composition-1K \cite{DIM} test set and the synthetic unified dataset. To explore how the user interactions perform on different types of data, we evaluate them on the synthetic unified dataset. The quantitative results are shown in Table \ref{tab:interactive_ablation_4classes}, while the qualitative visualization results are depicted in Figure \ref{fig:annotation_comparison}. Only using the foreground point is error-prone. Clicking on opaque region or transparent region is totally different, which may cause confusions between foreground prior and background prior. Scribble achieves better performance on SO and STM categories, attributing to much more foreground information. However, the results on NSO and NST images indicate it is not sufficient when faced with non-salient cases, limited by the lack of background hints. Foreground/background points and extreme points show better object locating ability on non-salient images, and the latter can provide much more stable prior, since it marks points from the boundary of object rather than from inside region. Given the information diversity and stability, and considering the labeling efficiency, bounding box is the best interactive type for the matting task, which gives an effective prior of coarse foreground region without errors and provides highly confident background hints. Moreover, bounding box can be acquired by 2 clicks, which is twice as fast as extreme points.

\begin{figure}[h]
\begin{center}
\includegraphics[width=\linewidth]{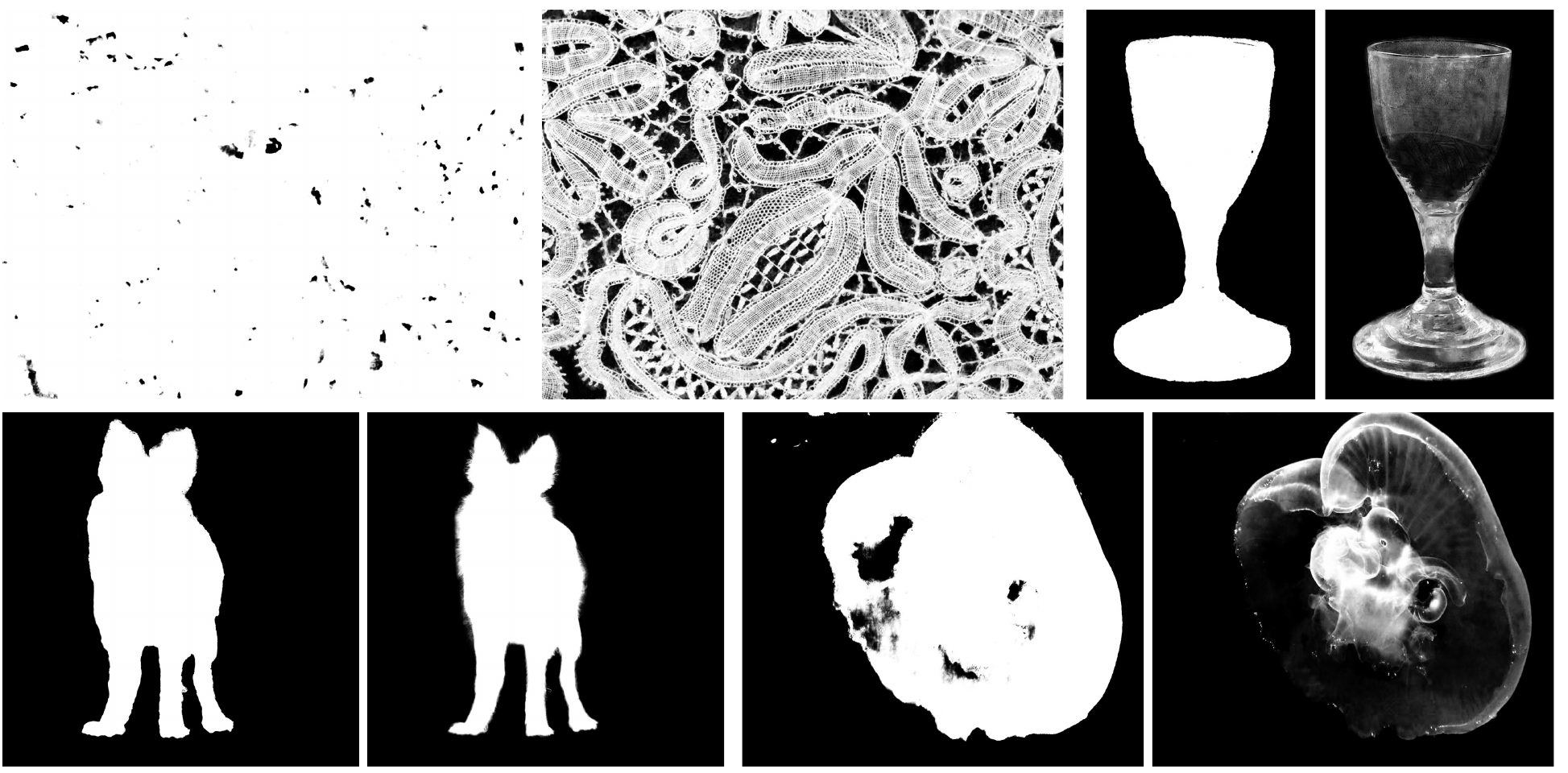}
\end{center}
  \caption{Visualization pairs of predicted mask and related alpha matte. The left of each pair is the predicted mask, while the right is the predicted alpha matte.}
\label{fig:mask2alpha}
\end{figure}

\begin{figure*}[t]
\begin{center}
\includegraphics[width=0.95\linewidth]{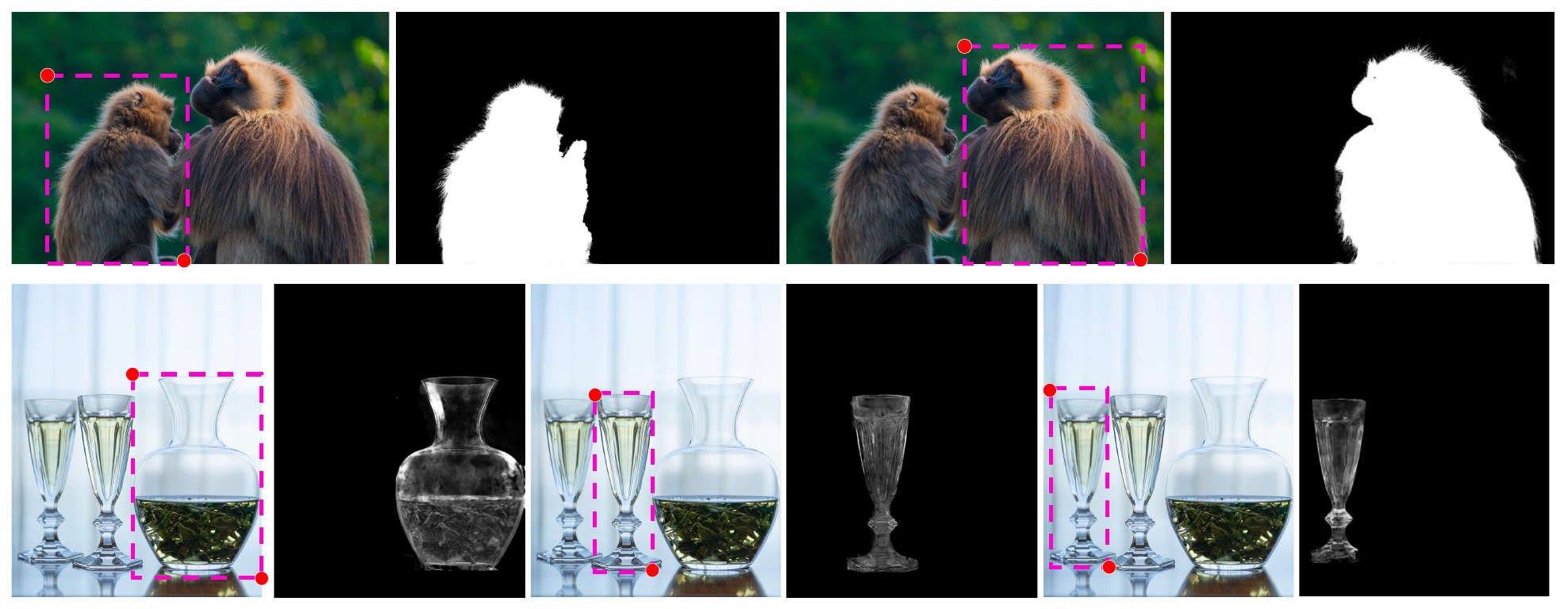}
\end{center}
  \caption{The visual results of UIM on real-world images. With specific bounding box interaction on object, UIM can predict precise alpha matte of the selected salient object.}
\label{fig:real_world}
\end{figure*}

\hspace*{\fill} \\
\noindent\textbf{Effectiveness of the components.} We evaluate three modules of UIM on Composition-1K test set. The baseline is the ResNet-34 based U-Net. And the results are presented in Table \ref{tab:module_ablation}, from which we can see that all the three modules can facilitate the matting task obviously and complement to each other. To valid the segmentation-matting decoupled design, we also evaluate the transparency mapping (TM) method on the synthetic unified dataset, which are shown in Table \ref{tab:tm_ablation}. Enhanced by the transparency mapping, samples from ST gain more improvement than that from SO, which indicates this method is helpful for transparent objects matting. Moreover, the decoupled design can provide better foreground region prediction, leading to better results on both opaque and transparent cases. In addition, the MAMF and FC methods further promote the alpha prediction quality.

To better explore the effectiveness of the components, we conduct a qualitatively analysis. 1) Transparency mapping. We visualize the predicted foreground mask and related alpha matte. As shown in Figure \ref{fig:mask2alpha}, the segmentation branch provides coarse foreground region prediction. For opaque objects, the matting branch refines the boundary and texture details. For transparent objects, the matting branch correctly maps the transparent information for each pixel. 2)MAFM. To better verify the mechanism of the proposed MAFM, we visualize the feature maps at the fourth layers of the matting decoder. As shown in Figure \ref{fig:mafm}, we can see that the details of the final predicted alpha matte are more obvious after MAFM processing, and some wrong or vague prediction are corrected by MAFM. For objects with different granularity of details, the module selects features with different weights, \emph{e.g.} high level features for main body information, and low level features for boundary and texture details.

 \begin{table}[h]
    \small
	\centering
	\caption{Module ablation studies on Composition-1K test set with bounding box interaction. The metrics are trimap-based. MAFM: multi-scale attentive fusion module; TM: transparency mapping design; FC: foreground consistency learning strategy.}	
	\begin{tabular}{ccc|cccc} 
	\toprule
     MAFM & TM & FC & MSE & SAD & Grad. & Conn.\\
	\midrule
    & & & 0.023 & 61.2 & 35.2 & 51.2 \\
    \Checkmark & & & 0.015 & 47.5 & 25.2 & 40.9 \\
    & \Checkmark  & & 0.019 & 57.6 & 33.7 & 49.9\\
    \Checkmark & \Checkmark  & & 0.014 & 42.6 & 21.7 & 37.4\\
    \Checkmark & \Checkmark  & \Checkmark & \textbf{0.012} & \textbf{38.2} & \textbf{17.9} & \textbf{33.8}\\
	\bottomrule
	\end{tabular}
	\label{tab:module_ablation}
\end{table}

 \begin{table}[h]
    \small
	\centering
	\caption{Ablation studies of transparency mapping method on the synthetic unified dataset. The metrics are trimap-free.}	
	\begin{tabular}{c|l|cccc} 
	\toprule
     Categories & Methods & MSE & SAD & Grad. & Conn.\\
 	\midrule
    \multirow{2}{*}{SO} & baseline & 0.015 & 17.0 & 9.0 & 14.0\\
     & +TM & \textbf{0.013} & \textbf{16.8} & \textbf{8.8} & \textbf{13.9}\\
	\midrule
    \multirow{2}{*}{ST} & baseline & 0.022 & 59.7 & 29.2 & 48.8\\
     & +TM & \textbf{0.018} & \textbf{54.7} & \textbf{27.3} & \textbf{47.8}\\
	\bottomrule
	\end{tabular}
	\label{tab:tm_ablation}
\end{table}

\subsection{Application Studies}

Interactive matting is an application-oriented task, which dedicates to matt on all types of images. To conduct a more objective evaluation of the proposed UIM, we explore its possibility in real world scenes and analyze the limitations.

\hspace*{\fill} \\
\noindent\textbf{Real-world image matting.} We use free-license images from real world scenes for test, with the model trained on Composition-1K. We manually generate bounding box interaction as auxiliary input, and the results are shown in Figure \ref{fig:real_world}. We can see that UIM can effectively predict matte of specific objects with related bounding box interactions, which is capable of handling both NSO and NST images. Results demonstrate that UIM can be applied to real-world image matting task, rather than being limited to the image under certain conditions (\emph{e.g.} salient foreground object and simple background). 

\begin{figure}[htbp]
\begin{center}
\includegraphics[width=\linewidth]{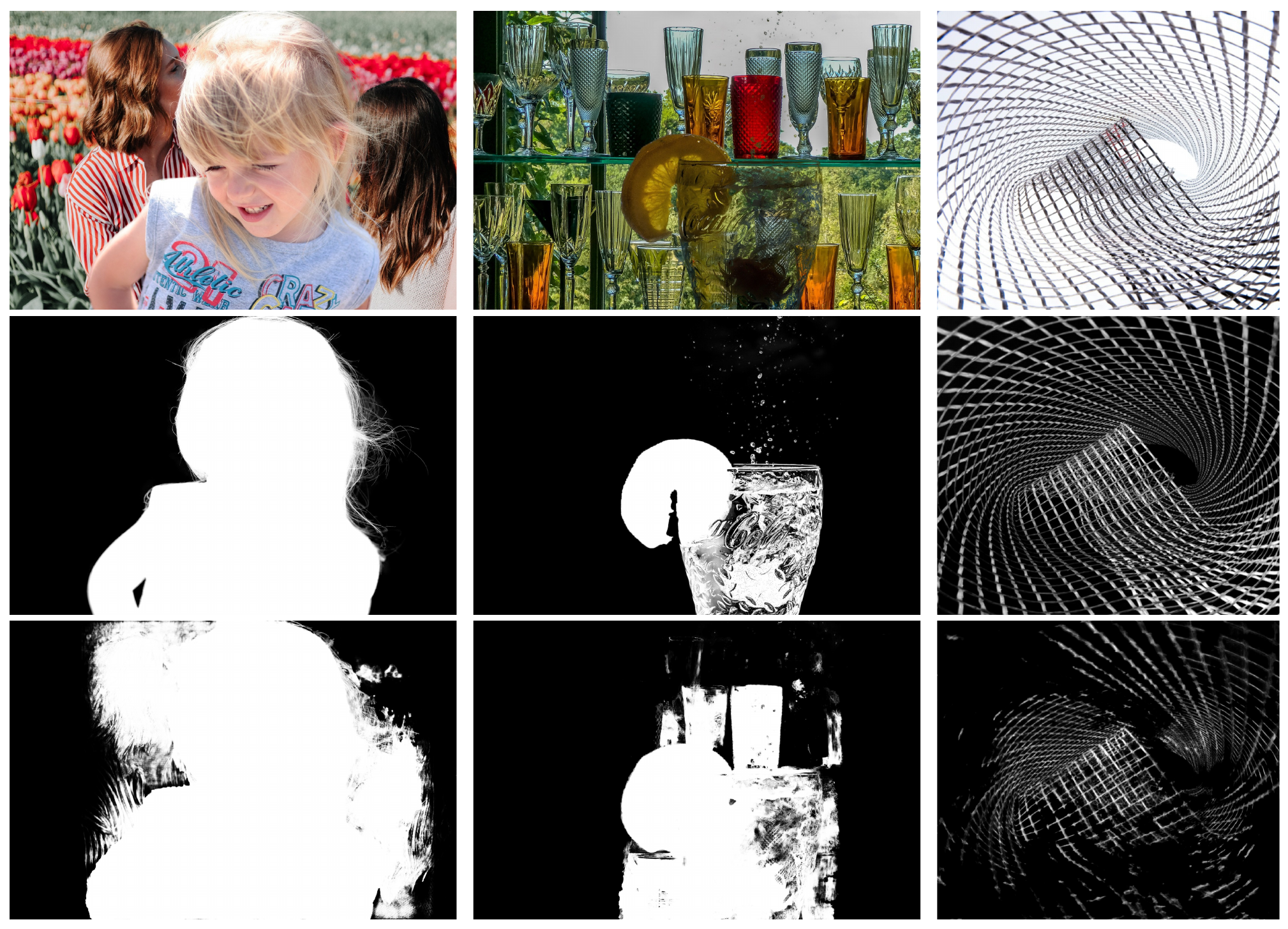}
\end{center}
  \caption{The failure cases of UIM on the synthetic unified dataset. The first row is the image, the middle row is the ground truth alpha, and the last row is the predicted alpha.}
\label{fig:failure_cases}
\end{figure}

\hspace*{\fill} \\
\noindent\textbf{Limitation analysis.} We analyze the failure cases on both synthetic datasets and real-world images. The qualitative results are illustrated in Figure \ref{fig:failure_cases}. First, bounding box interaction can only provide coarse prior of foreground, which is not sufficient for the complex and overlapped foregrounds in some cases. As shown in the first two columns, within the box area, there are multiple foregrounds with similar texture and color, leading to a failure prediction. Second, as shown in the third column, it is hard to make a precise prediction in the cases with frequently interlaced foreground and background, which remains to be a problem for all matting methods.

\section{Conclusion}

In this paper, we proposed UIM, a novel method for unified interactive image matting. UIM can leverage various user interactions to determine the matting target, effectively solving the ambiguity of multiple objects on a single image. To unify the matting performance of opaque and transparent objects, the matting task is decoupled into a foreground segmentation stage and a transparency prediction stage. 
Moreover, a new training strategy is designed for the synthetic image matting dataset to facilitate the capability of feature representation.
Experimental results demonstrate that the performance of UIM is superior to existing trimap-free and interactive matting methods, which is even competitive compared with the trimap-based methods. We believe that our proposed method can serve as a new paradigm for general image matting, and an effective solution for practical matting applications.




\bibliographystyle{elsarticle-num-names} 
\bibliography{ref_v1.bib}





\end{document}